# Understanding Deep Learning defenses Against Adversarial Examples Through Visualizations for Dynamic Risk Assessment

Xabier Echeberria-Barrio · Amaia Gil-Lerchundi · Jon Egana-Zubia · Raul Orduna-Urrutia

**Abstract** In recent years, Deep Neural Network models have been developed in different fields, where they have brought many advances. However, they have also started to be used in tasks where risk is critical. A misdiagnosis of these models can lead to serious accidents or even death. This concern has led to an interest among researchers to study possible attacks on these models, discovering a long list of vulnerabilities, from which every model should be defended. The adversarial example attack is a widely known attack among researchers, who have developed several defenses to avoid such a threat. However, these defenses are as opaque as a deep neural network model, how they work is still unknown. This is why visualizing how they change the behavior of the target model is interesting in order to understand more precisely how the performance of the defended model is being modified. For this work, some defenses, against adversarial example attack, have been selected in order to visualize the behavior modification of each of them in the defended model. Adversarial training, dimensionality reduction and prediction similarity were the selected defenses, which have been developed using a model composed by convolution neural network layers and dense neural network layers. In each defense, the behavior of the original model has been compared with the behavior of the defended model, representing the target model by a graph in a visualization.

**Keywords** Adversarial attacks · Adversarial defenses · Visualization

Xabier Echeberria-Barrio · Amaia Gil-Lerchundi · Jon Egana-Zubia · Raul Orduna-Urrutia
Vicomtech Foundation, Basque Research and Technology Alliance (BRTA), Mikeletegi 57, 20009 Donostia-San Sebastiá´n (Spain)
Tel.: +34943309230
Fax: +34943309230
E-mail: xetxeberria@vicomtech.org, agil@vicomtech.org, jegana@vicomtech.org and roduna@vicomtech.org

## 1 Introduction

Due to deep neural networks' computing power requirements, it was not until recently that they started being extensively studied and implemented in many fields which have a direct impact in humans' lives. Some of these fields, such as healthcare [9] and autonomous vehicles [23], are critical due to any misdiagnosis or decision making error could potentially lead to major incident which could compromise people's lives. Because of that, researchers have studied possible attacks to deep neural networks models discovering several threats on them. Once these vulnerabilities were studied, several defenses were proposed in order to create a model that is more robust to the attacks.

The widely known adversarial example attack has been extensively studied and attempts have been made to develop different defenses to make models more robust against them [1]. Adversaries take advantage of the sensitivity of target model by and adding a specifically designed noise to an input sample. Even if this distortion is imperceptible for humans, it is capable of modifying the original output prediction of the sample. The first adversarial example in deep neural network was generated by the algorithm L-BFGS [24]. Since then, more efficient and less detectable algorithms have been developed for adversarial example generation, such as Fast Gradient Sing Method (FGSM) [16], Basic Iterative Method (BIM) [14], Projected Gradient Descent Method (PGD), Jacobian-based Saliency Map Attack (JSMA) [20] and DeepFool [18]. Furthermore, some of the algorithms allow to generate adversarial attacks that are transferable from other models [6] and they are capable to generate them even with limited knowledge of the model [11].

To defend the system against adversarial attacks, the strategy pursued can be centered in constructing more robust models or adding external layers to the model to detect the attacks and block them.

This paper is the extension of the previous paper [8], which was presented in CISIS 2020 conference[1]. It presented two types of defenses called dimensionality reduction and prediction similarity. The dimensionality reduction defenses were based on previous defenses developed through autoencoders, but implemented new, more complex models. The prediction similarity defense was a new contribution to the methods for avoiding an adversary attack.

On one hand, the model can be modified to reduce its vulnerabilities but also maintaining the accuracy. For that purpose, adversarial train [24] introduces adversarial examples in the training set of the model so that the model learns to classify them correctly. Other possibility is to reduce the effect of the noise in the model by modifying its structure for example by adding autoenconders in the model structure or principal component analysis (PCA) algorithm to the input data. Autoencoders also can be applied separately in order to reconstruct input images adequately [10].

On the other hand, the target model remains untouched and the defense consist on adding another layer in the system such as an adversarial detector [15]. The detectors can be generated taking into account the input data properties and their effect in the model [17] or by trying to detect when certain delivered predictions are asked in order to construct the adversarial example [8]. One possibility is learning from the original internal behavior (such as activation values of neurons) of the model and how is this modified when adversarial examples are introduced. Other possibility is to detect when the adversary is trying to find the vulnerabilities of the model for generating the adversarial example, since usually multiple similar prediction responses are needed.

While the purpose of these defenses is clear, it is more difficult to know how their behavior is modified compared to the original model. Deep neural networks are an opaque machine, where decision making is really difficult to understand and that is the reason why the deep learning defenses are also opaque. Therefore, visualizing the behavior of the target model would be useful to understand the modification of the behavior when each defense is implemented.

Multiple visualizations have been proposed in order to understand what a deep neural network have learnt. Carter et al. [4] propose using activation atlases to be capable of visualizing multiple activation values of neurons of multiple images at once. Olah et al. [19] propose a block strategy where they visualize the most meaningful activation values at each spatial point of the image. With those blocks, they are capable to interpret the concepts that the neural network is creating in each part of the image. Kahng et al. [13] generated a visualization system with multiple coordinated views where the user is able to explore localized activation values together with model architecture and compare them for multiple instances. Selvaraju et al. [22] use gradient weighted activation mapping to generate a saliency map indicating the most meaningful parts of the image for each prediction class. These saliency maps can be interpreted as explanations indicating what the model is looking for in each image to detect which is the suitable class of the input image.

The rest of the paper is divided as follows: Section 2 explains the dataset and model selected for the experiments, adversarial attacks selected and studied defenses against them and the visualization technique used to understand their effect. The results are given in Section 3 and Section 4 lists the lessons that were learned and future work.

## 2 Methodology

This work studies three defenses: the widely known adversarial training defense and the previously proposed by the authors dimensionality reduction and prediction similarity defenses [8]. The contributions of previous paper were the implementations of two new defenses: the dimensionality reduction defense applied to a more complex deep learning model and the implementation of a new adversarial examples detector, called prediction similarity. Each defense and its particular implementation is detailed next. Finally, the visualization used for defense interpretation is described, which aims to represent the behavior of the analyzed model (the original and the defended version).

### 2.1 Dataset and original model to be protected

These defenses have been tested using breast cancer dataset[2], which is composed by two type of images: non-cancer images (class 0) and cancer images (class 1) [12, 5]. Moreover, a model composed by VGG16 and a dense neural network (DNN) layer. This structure is shown in Figure 1.

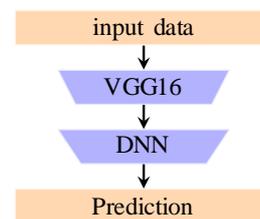

Fig. 1: Our model's structure. The VGG16 could be replaced by any CNN.

Developed defenses could easily be applied to other datasets and different deep learning structures as they do not need any specific type of layer or type of data.

---

[1] http://2020.cisisconference.eu/

[2] https://www.kaggle.com/paultimothymooney/breast-histopathology-images

## 2.2 Adverarial generation algorithms used

After searching for various adversarial attack methods in the literature, the following algorithms have been selected for the experiments due to their simplicity and efficiency.

- Fast Gradient Sing Method (FGSM) [7]
- Basic Iterative Method (BIM) [14]
- Projected Gradient Descent Method (PGD) [16]

The *foolbox* library[3] was used for the generation of the attacks.

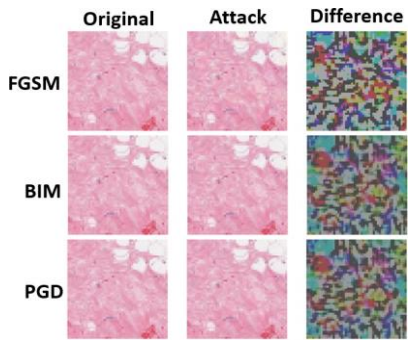

Fig. 2: Advesarial examples

For the experiments, the adversarial examples have been split in two groups, the known and the new adversarial examples. Known adversarial examples have been already computed before the defense is generated, which do not need further predictions to obtain them (they are already known). In contrast, new adversarial examples are those that are obtained once the defended model is implemented.

## 2.3 Defenses

### 2.3.1 Adversarial Training

The idea of adversarial training was introduced in [24]. Once the adversarial examples have been added to the training data, this defense retrains the targeted model with it, learning to classify them correctly. However, this technique only guarantees that the previously known adversarial examples introduced in the training data will be correctly classified, the new model does not achieve competitive robustness against new adversarial examples [25, 3].

Adversarial examples are obtained and then added to the original training data. Then, the new training data is used to retrain the DNN part of the model (it becomes the defended model, by adversarial training). Although the new model is more robust against the added adversarial examples, it is easy to obtain new ones and stay in an inexhaustible circle of attacking and defending.

---

[3] https://github.com/bethgelab/foolbox

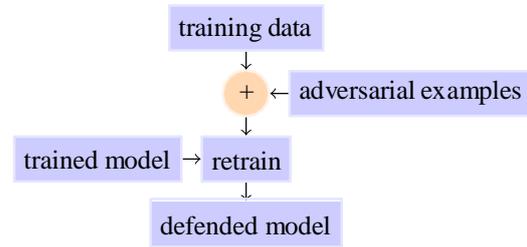

Fig. 3: Standard adversarial training.

This defense, is widely known and studied in the literature. For known adversarial examples, this defense trains the model to make it more robust against them, however when we talk about new adversarial examples, this defense does not make the target model robust. Adversarial training avoids known adversarial examples with a 90% (Tab.2) success rate, while new adversarial examples can be generated without much difficulty. With the new examples obtained, the model can be retrained again to ensure those vulnerabilities are also taken into account. Therefore, adversarial training is a never-ending defense.

### 2.3.2 Dimensionality Reduction

Different variants of this defense can be implemented, but all of them are based in the same idea. The used strategy is to add a dimensionality reduction layer (such as autoencoders) to remove non natural noise from the input sample. This extra process avoids adversarial examples, making the added noise in adversarial attack irrelevant. However, depending the place where the new dimensionality reduction layers are inserted, the robustness of the original model is different. The literature has studied the utility of these defenses in order to avoid adversarial examples [2]. In the particular case of deep learning models, there are multiple ways to reduce the dimensionality of data, such as adding CNNs and autoencoders layers [21, 10]. This subsection covers two variants of dimensionality reduction: middle autoencoder and initial autoencoder.

Once the outputs of data are obtained through CNN (VGG16 in our case), the **middle autoencoder** is trained using these outputs. After the autoencoder is trained, it is inserted before the DNN (Fig. 4), the CNN and DNN are maintained with the original structure (original weights). The base idea is that the middle autoencoder "cleans" the noise of CNN's outputs before using them as DNN's input data.

Otherwise, the **initial autoencoder** is inserted before the CNN, once the autoencoder is trained using the selected dataset. Again, both the CNN and DNN keep the original weights, since they are not retrained (Fig. 5). In this variant, the initial autoencoder "cleans" the image noise before making predictions with the initial model.

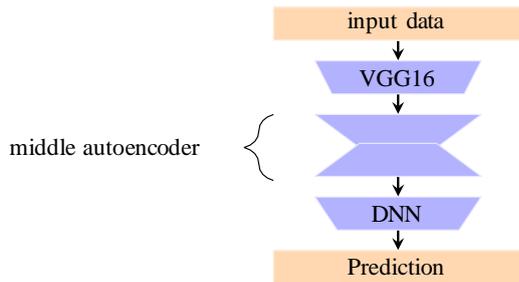

Fig. 4: Middle autoencoder model.

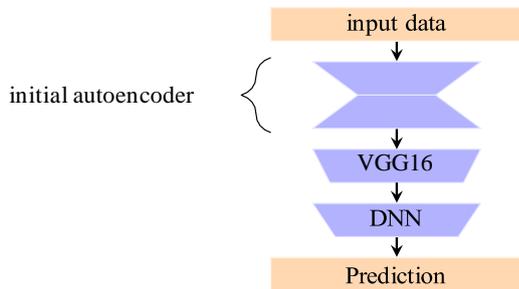

Fig. 5: Initial autoencoder model.

Even though both variants are based on the same idea to avoid adversarial examples, the results are different. Known adversarial examples are avoided with 60.4% success rate in middle autoencoder variant, while initial autoencoder defense avoid them with 70.5% success rate (Tab. 2). However, in the initial variant of the autoencoder, the accuracy of the original data has been lost, while the defense of the average autoencoder has maintained the accuracy of the original data similar to the original model (Tab. 1). Otherwise, both variants are more robust against new adversarial examples than adversarial training defense, as is visible in (Tab. 2), since some of the new adversarials become distinguishable for the human-eye. This means that it is more difficult to obtain new adversarial examples in the models which are defended with the dimensionality reduction defense.

*2.3.3 Prediction Similarity*

This defense does not modify the model directly, an external layer is added to the original model. This layer saves the history of inputs, predictions and specifically designed features. The features are inspired by the idea that adversarial attacks need several predictions of similar images to generate an adversarial example. From the data obtained in this layer, a risk assessment feature can be generated to evaluate if an input is adversarial or not. Similarity measures are used to compare the actual input data with the previous ones. If the risk value is high, i.e., if there is a high probability of being the actual image an adversarial example, this layer could take an action to avoid the adversarial attack.

User, image, prediction value (the class and the probability of this class), minimum distance (to all previous images), prediction alarm (number of times the percentage of the class is smaller) and distance alarm (number of images with distance less than threshold) have been selected as features saved in each prediction.

There are several algorithms to compute the similarity value between two images. The most widely used metrics are the mean squared error (MSE) and peak signal to noise ratio (PSNR). However, in the last three decades, different complex metrics have been developed inspired by the human vision perception [28]. Some of these metrics are structure similarity metric (SSIM) [30] and feature similarity metric (FSIM) [27].

Once the adversarial attack is detected, the action takes part to avoid this attack. In our case, it returns the opposite (or another) class, if the detector detects something suspicious. This makes the adversary believe that he/she has already achieved the adversarial example, when in fact it is not. However, there are more possible actions that the output layer could take, such as blocking or predicting with a secundary model.

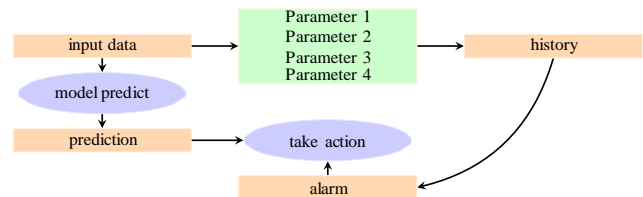

Fig. 6: Generalization of the prediction similarity defense.

Table 1: Results of the implemented defenses

| defenses | Original test data | Prediction impact |
|---|---|---|
| Without defense | 85.1% | |
| With adversarial training | 84.3% | Very low |
| With middle autoencoder | 82.4% | Low |
| With encoder | 82.1% | Low |
| With initial autoencoder | 70.0% | Medium |
| With prediction similarity | 85.1% | No impact |

This defense is focused on the detection of the process to get an adversarial example. That is why, the already obtained adversarial examples (i.e., known adversarial examples) are impossible to detect with this technique. However, the new adversarial examples, which need a process to be obtained, are detected with a 99.5% success rate (Tab. 2). Moreover, it does not modify the accuracy of the original test data, since this defense is only the addition of a detection layer which does not take part in model's prediction process, it makes its function in parallel.

Table 2: Results of the implemented defenses

| defenses | Known adversarials | New adversarials |
|---|---|---|
| Adversarial training | 90% | does not detect them. |
| Middle autoencoder | 60.4% | some of them distinguishable to the human eye. |
| Initial autoencoder | 70.5% | some of them distinguishable to the human eye. |
| Prediction similarity | 0% | detects the generation process 99.5% of the time. |

### 2.4 Visualization

These visualizations represent the behavior of the targeted model processing a input sample. In these graphs each vertex is a neuron from deep neural network and each edge is the connection between two neurons. The behavior has been represented though color in each vertex according to its relevance in the model's decision making. The mentioned behavior visualizations have been implemented only in model's dense part (770 neurons), due to the number of neurons in the complete model is 4694, that it would imply seven times more effort to study it completely. Moreover, it has to be taken into account that the VGG16 model is a pre-trained model and it is not specific of the resolution of the problem.

As it can be seen in the Figure 7, the visualizations are composed by three groups of nodes. Each set represents a layer from the dense neural network of the model. The nodes in the middle of the graph, which form a round set, represent the input neurons of the DNN part that are involved in the prediction of the input sample. The group, which are surrounding the input node group, represents the hidden layer. Finally, the set in the right side of the graph, which is composed by only two neurons, represents the output layer.

### 3 Experiment

The defenses from section 2 are designed as an countermeasure tool to avoid adversarial attacks. However, understanding the vulnerabilities of the model to be adversarial attacks effective has received little attention. The interpretability is a useful method to shed some light on the behavior of machine learning models. For this particular case, visualization is used as a explanation tool in order to study how the behavior of models are changed depending on the used defensed method. As mentioned above, the original model to be defended is composed of a convolutional neural network (VGG16) and a dense neural network (DNN). The interpretability is focused on the dense part of the model and the visualization is designed to observe the change on the behavior of the dense part depending on the input data received.

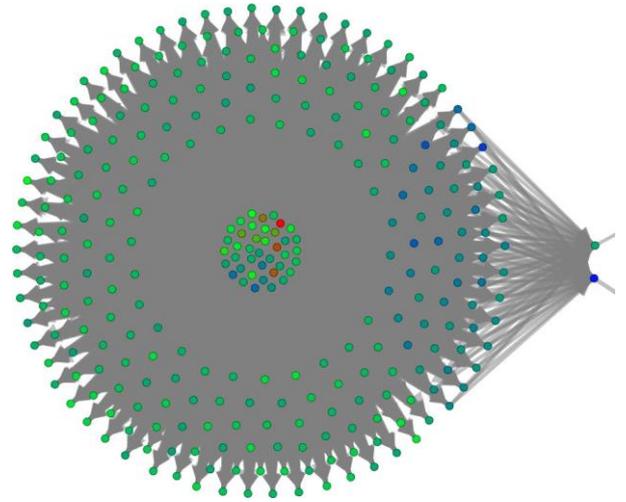

Fig. 7: Visualization example

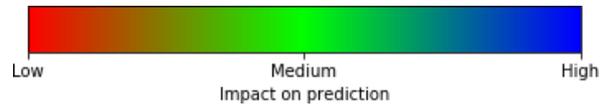

Fig. 8: The colormap which generate an assignation between the impact of a neuron in the model's prediction and a color

The structure of the dense part of the model is represented by a graph, where the vertices represent the neurons and the edges the relations between them. The color of each vertex represents the impact value of the specific neuron, where higher value implies a greater repercussion in the prediction process. The importance of each neuron (vertex) is computed calculating the difference between the output and inputs. The impact is calculated using the next equation:

$$Impact = Output - \sum_{i=1}^{n} Input_i \qquad (1)$$

where *Input* is the multiplication of the activation value of the previous neurons multiplied by the weight corresponding to the connection these two neurons. In this way, neurons that have a strong negative or positive influence can be identified, thus showing their impact on the prediction. In case the result is a negative value, it means that this neuron has a negative influence on the prediction and in the opposite case (a value greater than zero) it means that this neuron has a positive influence on the prediction.

In Figure 8 the colormap used in the graphs is shown, where the color represents the impact of the neurons using red for low, green for medium and blue for high impact values. The impact values belong to different scales depending on the sample used as input. Therefore, it is not possible to compare them between different behavioral graphs. However, it is possible to compare the number of neurons with low values and relevant values (calculated locally) between these graphs.

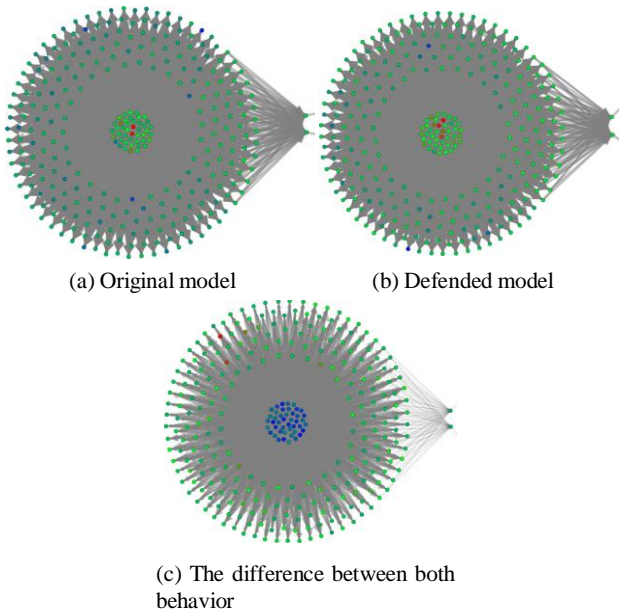

(a) Original model  (b) Defended model

(c) The difference between both behavior

Fig. 9: Behavior graphs generated through different types of imagery

ber of non-zero connection to output layer that have been modified in the hidden layer and how reliably the adversarial example has been learned. The figure 9c shows the non-zero connections from the hidden layer to the output layer that the defense has modified, since those are the connection differences between the original model (Fig. 9a) and defended model (Fig. 9b) using the same image. This difference in the number of connections is highlighted in the figure 10.

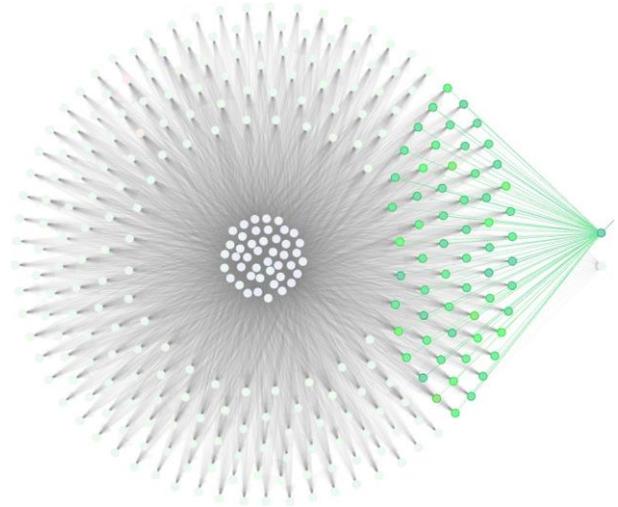

Fig. 10: Connections between hidden and output layers of the representation 9c

### 3.1 Adversarial Training

Adversarial training defense retrains DNN part of the original model with adversarial examples, resulting in the defended model. This is why the adversarial examples of the original model have a different behavior in the activations of the defended model.

In figure 9 three representation graphs can be observed, where figure 9a represent the behavior of the original model using an adversarial as input data, figure 9b shows the patterns generated by the same adversarial example in the defended model and figure 9c shows the difference between impact values of both cases.

Several details can be observed in these comparisons. On the one hand, it has been observed that this defense modifies the input layer significantly, while the effect in hidden layer is less remarkable. Thus, these visualizations show the importance of the input layer of the DNN, since VGG16 part have not been retrained and the image is the same, that is, the features which are input of the DNN part are the same in both cases.

On the other hand, in a study of 1000 adversarial examples, a correlation of 0.756 has been found between the num-

### 3.2 Dimensionality Reduction

Dimensionality reduction defense uses autoencoders to avoid the noise of the data as much as possible. Thus, this noise reduction generates a different behavior in the defended model when the adversarial examples of the original model are used as input. Moreover, the two different versions of the defended model are more robust against possible new adversarial examples. In other words, it is more difficult to obtain new adversarial examples from them, in some cases being adversarial examples even distinguishable to the human eye.

Those defenses do not modify the DNN, i.e., the DNN's weights are the same. Therefore, the thing that is changing is the input data of the dense neural network part. In the initial autoencoder defense, the input values of the CNN part are modified, whereas the inputs of the dense part are directly modified by the middle autoencoder defense.

In the figure 11 three graphs can be observed that represent the behavior of the different models (original and both defended) when the same adversarial example of the original model is used as input at each model. Note that the adversarial example used for graph generations is not an adversarial example in defended models. The first graph (figure 11a) shows the behavior of the original model, while the other two represent the behavior of the middle autoencoder model (figure 11c) and the initial autoencoder (figure 11b) respectively.

Observing these graphs, it can be noticed that each defense has a different impact on the behavior of the original model. One of the most remarkable changes when figures 11a and 11b are compared is the number of neurons in the input layer that take part in the prediction, which has decreased significantly (from 81 in the original model to 36 in the initial autoencoder defended model). Furthermore, the number of neurons from the hidden layer that take part in the final prediction increase (from 41 in the original model to 58 in the initial autoencoder defended model). In the middle autoencoder defense (figure 11c), these behavioral changes are not that evident. In the case of the number of neurons in the input layer, a little increase in the number can be observed (from 81 neurons in the original to 91 in the middle autoencoder defended model). By contrast, in the case of the hidden layer, this defense does not make any significant change (from 41 in the original model to 25 in the initial autoencoder defended model).

To study these modifications, for each adversarial example, the number of neurons that take part in its prediction is counted for each layer of the original model and each defended model. Once the neurons have been counted, the quantities of the input layer of the original model are compared with those of the input layers of the defended models, calculating the difference between them. This process is repeated with the quantities obtained in the hidden layer. Thus, four different values are obtained for each adversarial example: difference between the number of neurons in the input and hidden layer participating in the prediction of the original model and the initial autoencoder defended model and difference between the number of neurons in the input and hidden layer participating in the prediction of the original model and the middle autoencoder defended model. These values for all the adversarial images are shown in figure 12.

In the case of the input layer of the initial autoencoder defense almost all the differences obtained are positive (shown in figure 12a), which indicates that in almost all cases the number of neurons participating in the prediction has reduced. However, in the hidden layer comparing the original model with the same defense, almost all the differences obtained are negative (see figure 12c), i.e,. the number of neurons taking part in the prediction has increased. For the middle autoencoder defense is clear that most of the differences obtained from the input layer are negative (shown in figure 12b), i.e., the number of neurons participating in the prediction has increased mostly. However, for the hidden layer nothing can be concluded (see figure 12d), since there is not any clear trend.

### 3.3 Prediction Similarity

The prediction similarity defense generates a detector, in other words, it does not modify the behavior of the original model. Therefore, the behavior graphs of the original model and the defended model (original model plus detector) are the same. This defense can detect an adversarial example search process from an original image. In this case, we are interested in how the graph's behavior changes during the multiple iterations with the model needed to generate the adversarial example.

In the process of getting the adversarial example, several similar images are used as input successively. The figure 15 shows inputs in different iterations of this process, starting from the first graph (figure 15a) that represents the behavior of the original image to the final graph (figure 15e) showing the behavior of the obtained adversarial example. Furthermore, the figure 16 represent the difference between the original image behavior graph and the behaviour graph of each step in 15. In this particular case, the adversarial was detected at the step 12, but the adversarial example searching process was not stopped until the adversarial example was obtained (step 100) in order to study the detection process in all the possible steps.

In the current example, the SSIM distance between an image from a particular step and the one in the previous step is around 0.94. Taking into account that the distance (using the same metric) between two images of the dataset

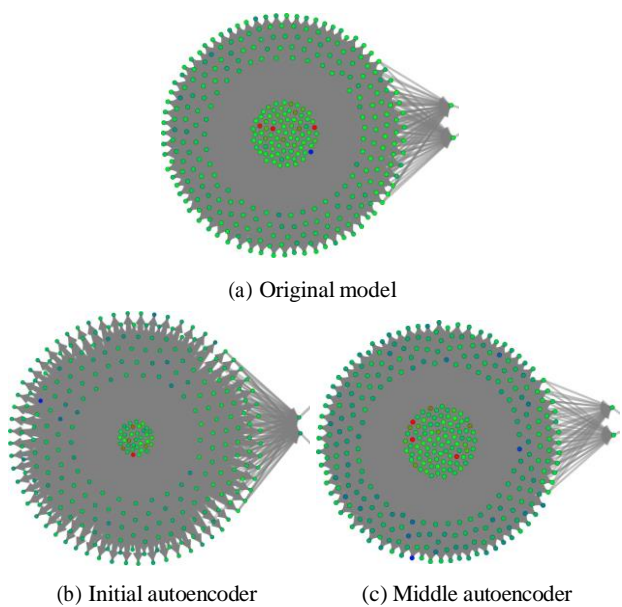

(a) Original model

(b) Initial autoencoder    (c) Middle autoencoder

Fig. 11: Behavior graphs generated through the same adversarial example

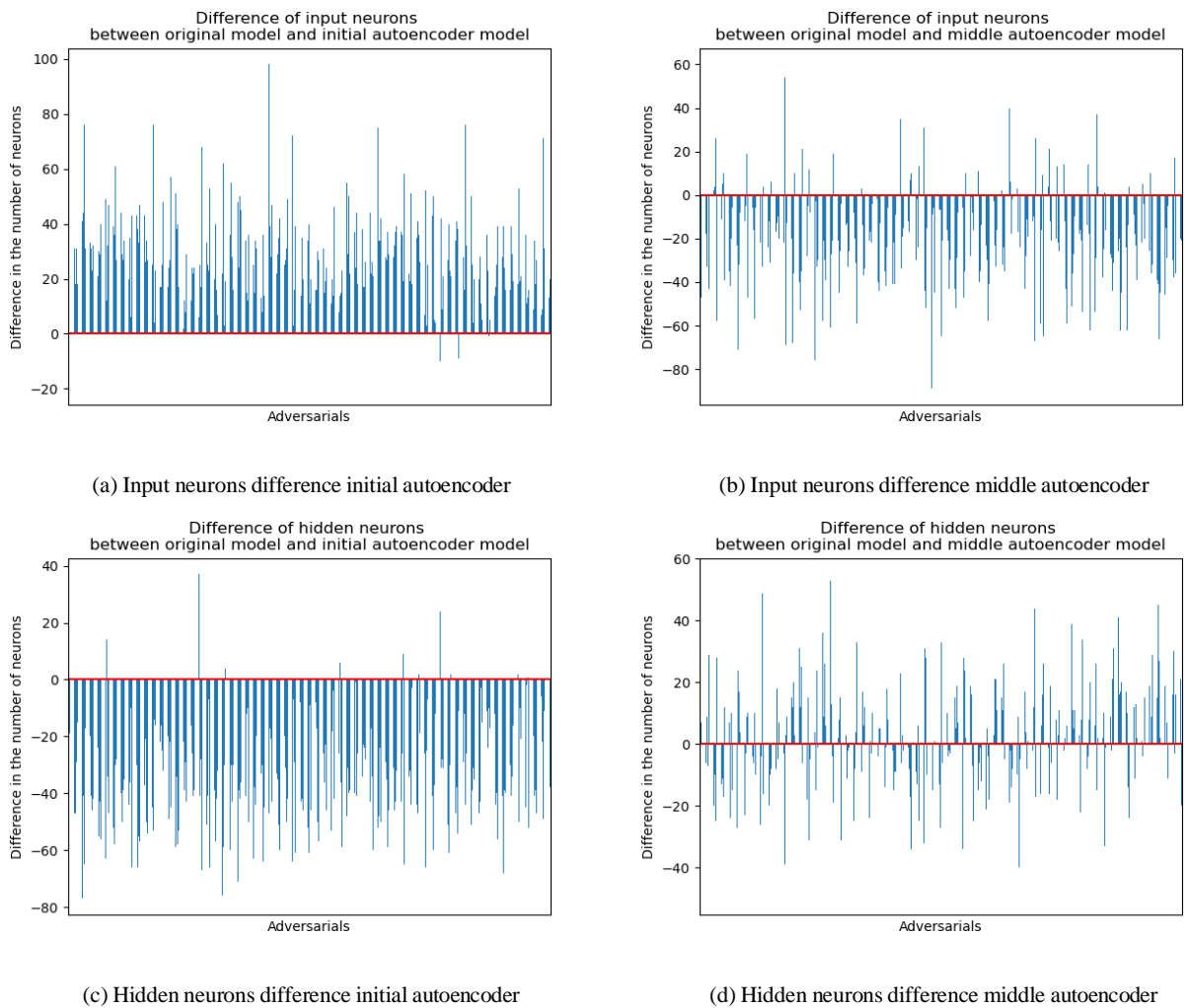

(a) Input neurons difference initial autoencoder

(b) Input neurons difference middle autoencoder

(c) Hidden neurons difference initial autoencoder

(d) Hidden neurons difference middle autoencoder

Fig. 12: Differences between the number of neurons in the original model and in each defended model, separated by layers for all the adversarial examples in the dataset.

is around 90, the change generated by this adversarial example generation algorithm between two consecutive steps is minimal. However, the figure 16a shows that this small change already has a considerable impact on the behavior of the neurons.

Moreover, observing the differences with the original image during the process (shown in figure 16), it can be concluded that the neurons that get more affected by change the ones from the input layer of the DNN. Therefore, an analysis of the input neurons of the DNN has been carried out to observe their evolution in the process of an adversarial example acquisition. A total of 69381 test images[4] were used to study which neurons participated significantly in the prediction for a particular image class. A neuron participates in the prediction of an image if the assigned value in the behavior visualization is non-zero. For this purpose, the participation of each neuron per class was calculated counting the number times a neuron participated, i.e., the number of times the activation value different from zero. This value ob-tained per class has been rescaled using the total number of images in each class and refer to as frequency. As this dataset consists of only two classes, two frequencies would be obtained for each neuron: the frequency with respect to class 0 and the frequency with respect to class 1. The difference of both frequencies has been calculated to study which of the neurons have more polarized behavior towards one of the classes. After following this process, several neurons were detected with a significant participation towards one of the classes and their frequency values per class and the differences between them are shown in table 3. The evolution in the behavior of these identified neurons has been studied separately for each class. As expected, in the case of the processes of an image of class 0, these behaviors show a decrease in the impact of the neurons associated with this class throughout the process, i.e., the absolute values of the activations of the most meaningful neurons get reduced to obtain an image of class 1 and an increase in the impact of the neurons associated with the opposite class. While in the processes of generating an adversarial example for a image from the class 1 the opposite evolution can be observed. The aforementioned process can be visualised for an image in each of the two classes in figure 13. Thus, in the difference between the adversarial example obtained and the original image, which belongs to class 0, have several neurons associated with this class colored in blue (figure 14), indicating that these neurons have experienced major modifications with respect to the rest of the neurons.

[4] https://www.kaggle.com/paultimothymooney/breast-histopathology-images

Table 3: Neuron frequencies in prediction participation by class

| Neuron | Class 0 frequency | Class 1 frequency | Frequency difference |
|---|---|---|---|
| 28  | 0.7279 | 0.1958 | 0.5320 |
| 226 | 0.8659 | 0.3365 | 0.5293 |
| 44  | 0.5916 | 0.1438 | 0.4477 |
| 486 | 0.8349 | 0.4002 | 0.4346 |
| 124 | 0.5274 | 0.1325 | 0.3948 |
| 435 | 0.4329 | 0.0568 | 0.3760 |
| 254 | 0.2820 | 0.6419 | 0.3599 |
| 40  | 0.0700 | 0.4210 | 0.3510 |
| 265 | 0.1691 | 0.4804 | 0.3112 |
| 76  | 0.1486 | 0.4517 | 0.3030 |

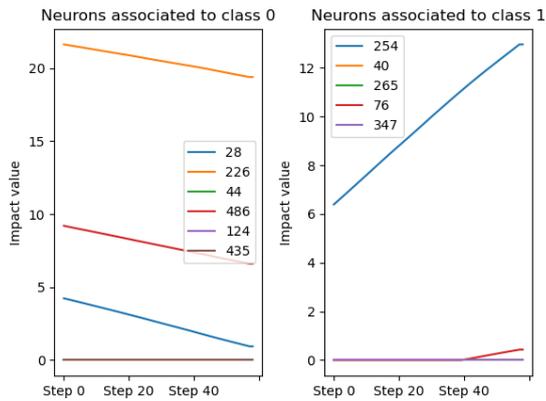

Fig. 13: Evolution of significant neurons associated to each class in the adversarial example generation process of example shown in 16.

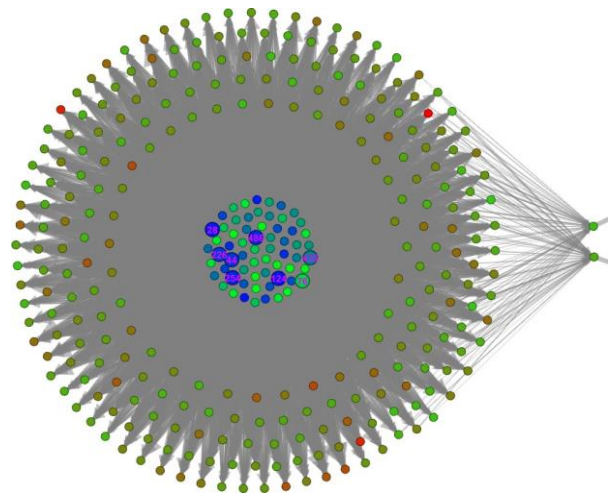

Fig. 14: Change of behavior of neurons associated with a particular class (shown in table 3) in the adversarial example generation process (Fig. 15). The size of indicated neurons are amplified and the neurons are labeled with their number in purple.

Therefore, it can be observed how a set of neurons has a relevant importance in the prediction of a class of images, and these neurons to be studied can be reduced thanks to these visualizations in order to make a more focused analysis on the part of the model that is of interest. This allows the development of a more optimal detector by focusing on a smaller number of neurons, these being those associated with a class.

## 4 Conclusions and future work

This work presents how different defenses against adversarial attacks modify the behavior of the model. Using visualizations, behaviors of the target model in different implementations have been shown in order to gain a deeper understanding of these defenses. In the three studied defenses, changes in the behavior of the model have been observed, which can be useful for improving and optimizing these defenses. The observed features shed light on the opacity of deep neural networks and how the defenses generated modify the internals of the model to counter these vulnerabilities. In addition, these generated visualizations and the results obtained can be implemented in other models with other datasets.

In the future, this knowledge will allow to develop more efficient defenses and detectors to combat different threats, including the adversary attack, using this visualizations. Moreover, it will be possible to develop detectors that work directly on these behavioral graphs, generating a new type of detectors [29, 26] implemented on deep neural networks.

**Acknowledgements** This work is funded under the SPARTA project, which has received funding from the European Union Horizon 2020 research and innovation programme under grant agreement No 830892.

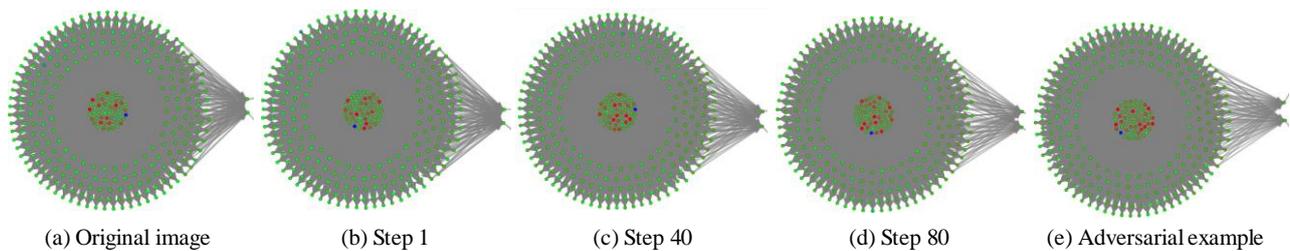

(a) Original image  (b) Step 1  (c) Step 40  (d) Step 80  (e) Adversarial example

Fig. 15: Process of obtaining an adversarial

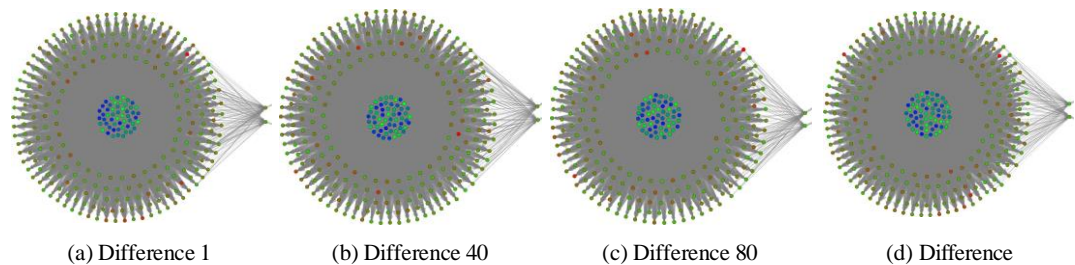

(a) Difference 1  (b) Difference 40  (c) Difference 80  (d) Difference

Fig. 16: Differences between original image behavior graph (Fig. 15a) and the other behavior graphs (Fig. 15b, Fig. 15c, Fig. 15d, Fig. 15e)

**Conflict of interest**

The authors declare that they have no conflict of interest.